\documentclass[a4paper]{spie}  

\usepackage{amsmath, amssymb}

\usepackage{amsfonts}
\usepackage{graphicx}
\usepackage{cite}
\usepackage{textcomp}
\usepackage{rotating}

\newcommand{\vc}[1]{{\bf #1}}

\newcommand{\ma}[1]{{\bf #1}}

\title{Laplacian embedding with tuned localization and spectral bandwidth using graph Slepians}

\title{Guiding Network Analysis using Graph Slepians:\\ An Illustration for the {C. Elegans} Connectome}

\author{Dimitri Van De Ville$^{1,2}$, Robin Demesmaeker$^1$, Maria Giulia Preti$^{1,2}$
\skiplinehalf
$^1$ \'Ecole Polytechnique F\'ed\'erale de Lausanne (EPFL), Switzerland\newline
$^2$ University of Geneva, Switzerland
}

\authorinfo{Further author information: 
(dimitri.vandeville, robin.demesmaeker, maria.preti)@epfl.ch}

\begin{document}
\maketitle

\begin{abstract}
Spectral approaches of network analysis heavily rely upon the eigendecomposition of the graph Laplacian. For instance, in graph signal processing, the Laplacian eigendecomposition is used to define the graph Fourier transform and then transpose signal processing operations to graphs by implementing them in the spectral domain. Here, we build on recent work that generalized Slepian functions to the graph setting. In particular, graph Slepians are band-limited graph signals with maximal energy concentration in a given subgraph. We show how this approach can be used to guide network analysis; i.e., we propose a visualization that reveals network organization of a subgraph, but while striking a balance with global network structure. These developments are illustrated for the structural connectome of the C. Elegans. 
\end{abstract}

\keywords{Graph signal processing, Graph Laplacian, Slepian functions, Graph cut, Laplacian embedding}

\section{Introduction}
\label{sec:intro}
In many applications, data can be advantageously modeled and analyzed using graphs, in particular, to reveal emergent, global network properties. In addition, increased availability of large and rich datasets, combined with better computational resources, have been driving the growing interest in network science and graph signal processing, respectively by different communities. One fundamental concept in network analysis is graph partitioning and embedding; e.g., how nodes can be grouped in subgraphs or how  they can be represented in a low-dimensional space based on the graph topology~\cite{Luxburg.2007}. Spectral graph theory relates these problems to the graph Laplacian and its eigendecomposition~\cite{Chung.1997}. For instance, the convex relaxation of the graph cut can be solved by thresholding the Laplacian eigenvector with the smallest non-zero eigenvalue---known as the Fiedler vector~\cite{Fiedler.1989}. Moreover, the Laplacian eigenvectors also provide a meaningful embedding by mapping nodes onto a line (or higher-dimensional space) that minimizes distance between connected nodes~\cite{Belkin.2003}. For graph signal processing, the eigenvectors of the graph Laplacian can be used to define a graph Fourier transform (GFT) and then to define processing steps in the spectral domain~\cite{Shuman.2013}; e.g., the spectral graph wavelet transform~\cite{Hammond.2011} is such a notable example. 

Recently, we have introduced the graph equivalent of Slepian functions~\cite{vandeville1701} that were proposed fifty years ago on regular domains to find a trade-off between temporal and spectral energy concentration~\cite{Slepian.1961,Slepian.1978}. The design of graph Slepians is based on an optimization criterion that expresses energy concentration of a band-limited graph signal in a predefined subgraph. In our previous work~\cite{vandeville1701}, we proposed to generalize this criterion such that it reflects a modified Laplacian embedding distance, which basically gives a meaning to eigenvalues as ``localized frequencies'', instead of as energy concentrations. 

Here, we apply the graph Slepian framework to guide network analysis. After a brief introduction on the theory and design of graph Slepians, we show how the information revealed by graph Slepians can be summarized to provide insightful visualizations. We then focus on the structural connectome of the \emph{C. Elegans} organism as an example to guide network analysis. 

\section{Graph Slepians in a Nutshell}

\subsection{Graph Fourier Transform}
Let us consider an undirected weighted graph with $N$ nodes. Its connectivity is characterized by an $N\times N$ symmetric adjacency matrix $\ma A$, where the elements $A_{i,j}$ indicate the edge weights between nodes $i,j=1,\ldots,N$. We assume the graph is connected and the edge weights are positive. Next to the graph and its topology, we can also consider graph signals that define a mapping from the nodes to a length-$N$ vector associating a value with every node; e.g., the graph topology models the presence of communication links between nodes, while the graph signal represents measures taken at the nodes. 

The graph Laplacian is defined as $\ma L'=\ma D-\ma A$, where $\ma D$ is the diagonal degree matrix with elements $D_{i,i}=\sum_{j=1}^N A_{i,j}$. We further consider the normalized graph Laplacian $\ma{L}=\ma{D}^{-1/2}\ma{L}'\ma{D}^{-1/2}$ that factors out differences in degree and thus is only reflecting relative connectivity.

Spectral methods for graphs are based on the insight that the eigendecomposition of the graph Laplacian gives access to a GFT\cite{Shuman.2013}. The eigenvectors of $\ma L$ minimize 
\begin{equation}
  \label{eq:lapl}
  \lambda = \frac{\vc u^T \ma L\vc u}{\vc u^T \vc u}
\end{equation}
and can be ordered according to increasing eigenvalues ${\lambda_1=0\le \lambda_2 \le \ldots \le \lambda_N}$. The eigenvectors play the role of basis vectors of the graph spectrum, and the associated eigenvalues the one of frequencies~\cite{Chung.1997}. 
By stacking the vectors $\vc u_k$ as the columns of $\ma U$, which is an $N\times N$ unitary matrix, we obtain the GFT that links a graph signal $\vc s$ and its spectral coefficient vector $\hat{\vc s}$ as 
$$
   \vc s = \ma U \hat{\vc s}, \text{ and } \hat{\vc s} = \ma U^T \vc s.
$$

\subsection{Graph Slepians}
To generalize Slepians to graphs, we introduce selectivity and bandwidth for the graph signals~\cite{Tsitsvero.2016,vandeville1701}. First, selectivity can be specified by a subset $\mathcal{S}$ that contains the $N_S$ nodes in which we want the energy concentration to be optimal. We introduce the $N\times N$ diagonal selection matrix $\ma{S}$ with elements $S_{k,k}=0/1$ that indicate the presence of index $k$ in $\mathcal{S}$; so the number of selected indices is $\text{trace}(\ma{S})=N_S$. 
Second, for the notion of ``bandwidth'', we propose to restrict the spectrum to the eigenvectors with the $N_W<N$ smallest eigenvalues $\lambda$. We then consider the truncated graph spectrum matrix $\ma U_W$ of size $N\times N_W$; i.e., it only contains the first $N_W$ columns of $\ma U$. Consequently, any band-limited graph signal can be represented by $$\vc{g}=\ma{U}_W \hat{\vc{g}}.$$

Finding the linear combination of eigenvectors within the bandlimit $N_W$ and with maximal energy in $\mathcal{S}$ reverts to optimizing the Rayleigh quotient 
\begin{equation}
\label{eq:slepian}
  \mu = \frac{\hat{\vc g}^T \ma{U}^T_W \ma{S} \ma{U}_W \hat{\vc g}}{\hat{\vc g}^T \hat{\vc g}},
\end{equation}
where $\ma{C}= \ma{U}^T_W \ma{S} \ma{U}_W$ is the concentration matrix. Since $\ma{L}$ is real and symmetric, $\ma{U}$ is real as well and we revert to the regular transpose $\cdot^T$.
The Slepian vectors are orthonormal over the entire graph as well as orthogonal over the subset $\mathcal{S}$; i.e., we have $\vc{g}_k^T \vc{g}_l=\delta_{k-l}$ as well as $\vc{g}_k^T \ma{S} \vc{g}_l  = \mu_k \delta_{k-l}$, where $\delta$ is the Kronecker delta.

By convention, the eigenvalues of the Slepian eigendecomposition are sorted according to decreasing energy concentration ${1>\mu_1\ge \mu_2 \ge \ldots > 0}$ in $\mathcal{S}$. The eigenvalue spectrum presents a sudden transition between well-localized and poorly-localized eigenvectors.  

\subsection{Graph Slepians Revisited: The Link with Laplacian Embedding}
The GFT plays a central role in Laplacian embedding and graph clustering. Specifically, in Laplacian embedding, the aim is to find a mapping from the nodes onto a line such that strongly connected nodes stay as close as possible, which can be expressed as~\cite{Belkin.2003}:
\begin{equation}
  \label{eq:lap_embedding}
  \arg \min_{\vc g} \sum_{i,j=1}^{N} A_{i,j} (g_i - g_j)^2 = \arg \min_{\vc g} \vc{g}^T \ma{L} \vc{g},
\end{equation}
where $\vc{g}^T\vc{g}=1$ and $\vc{g}^T \vc {1}=0$. The solution is the eigenvector of the Laplacian with the smallest non-zero eigenvalue; i.e., the Fiedler vector~\cite{Fiedler.1989}. The optimization (\ref{eq:lap_embedding}) is also related to the classical graph cut  problem that consists of partitioning the graph into clusters of nodes such that the cut size is minimal. In this setting, $g_i=\pm 1$ indicate the labels of the nodes, which can be relaxed to take any real values. 

We now use the Slepian design to generalize the criterion~(\ref{eq:lap_embedding}) as to find a bandlimited solution that is optimized for a selected set of nodes. To reveal this link, we first rewrite the quadratic form (\ref{eq:lap_embedding}) as
$$
  \vc{g}^T \ma{L} \vc{g} = \vc{g}^T \ma{U} \ma{\Lambda} \ma{U}^T \vc{g} = \hat{\vc g}^T \ma{\Lambda} \hat{\vc g}=\hat{\vc g}^T \ma{\Lambda}^{1/2} \ma{U}^T \ma{U} \ma{\Lambda}^{1/2} \hat{\vc g}.
$$
Therefore, we propose  to obtain a generalized Slepian criterion by introducing in this quadratic form bandwidth and node selection in the spectral and graph domains, respectively: 
\begin{equation}
  \label{eq:slepian-emb}
  \xi= \frac{\hat{\vc g}^T \ma{\Lambda}^{1/2}_W \overbrace{\ma{U}^T_W \ma{S} \ma{U}_W}^{\ma{C}} \ma{\Lambda}^{1/2}_W \hat{\vc g}}{\hat{\vc g}^T \hat{\vc g}},
\end{equation}
{where $\ma{\Lambda}_W$ is the band-limited $N_W\times N_W$ diagonal matrix.} 
{Consequently, we propose to solve the eigendecomposition of the modified concentration matrix $\ma{C}_\text{emb} = \ma{\Lambda}^{1/2}_W \ma{C} \ma{\Lambda}^{1/2}_W$.}
As before, the Slepian vectors $\vc g_k = \ma U_W \hat{\vc g}_k$ are double orthogonal as we have $\vc{g}_k^T \vc{g}_l=\delta_{k-l}$ and $\vc{g}_k^T \ma{S} \vc{g}_l=\xi_k \delta_{k-l}$.
We store the graph Slepians as columns of the matrix $\ma G$ of size $N\times N_W$. The concentration value of each graph Slepian can still be recovered using the quadratic form $\mu=\hat{\vc g}^T \ma C \hat{\vc g}$, or, vice versa, the modified embedded distance can be found for the previous design as $\xi=\hat{\vc g}^ \ma C_\text{emb} \hat{\vc g}$.  

This demonstrates that Laplacian embedding can be generalized as a Slepian problem with the additional weighting of the Laplacian eigenvalues. It is interesting to note that the eigenvalues $\xi$ of $\ma{C}_\text{emb}$ represent the modified embedded distance, or, equivalently, a ``frequency'' that is localized in the subgraph $\mathcal{S}$, whereas the eigenvalues $\mu$ of $\ma{C}$ represent the energy concentration in the subgraph.

\section{An Illustration for the Worm Connectome} 
The C.~Elegans is an intensely studied organism in biology. In particular, the wiring diagram of its 302 neurons has been carefully mapped during a long and effortful study~\cite{White.1986}. Here, we use the adjacency matrix that considered 279 somatic neurons and combined connectivity from chemical synapses and gap junctions~\cite{Chen.2006}. We retrieved the type of each neuron (including polymodal status) from the WormAtlas\footnote{http://www.wormatlas.org/}. 

\subsection{Graph Laplacian Embedding}
\begin{sidewaysfigure}[p]
\centering
\begin{tabular}{c}
\includegraphics[width=\textheight]{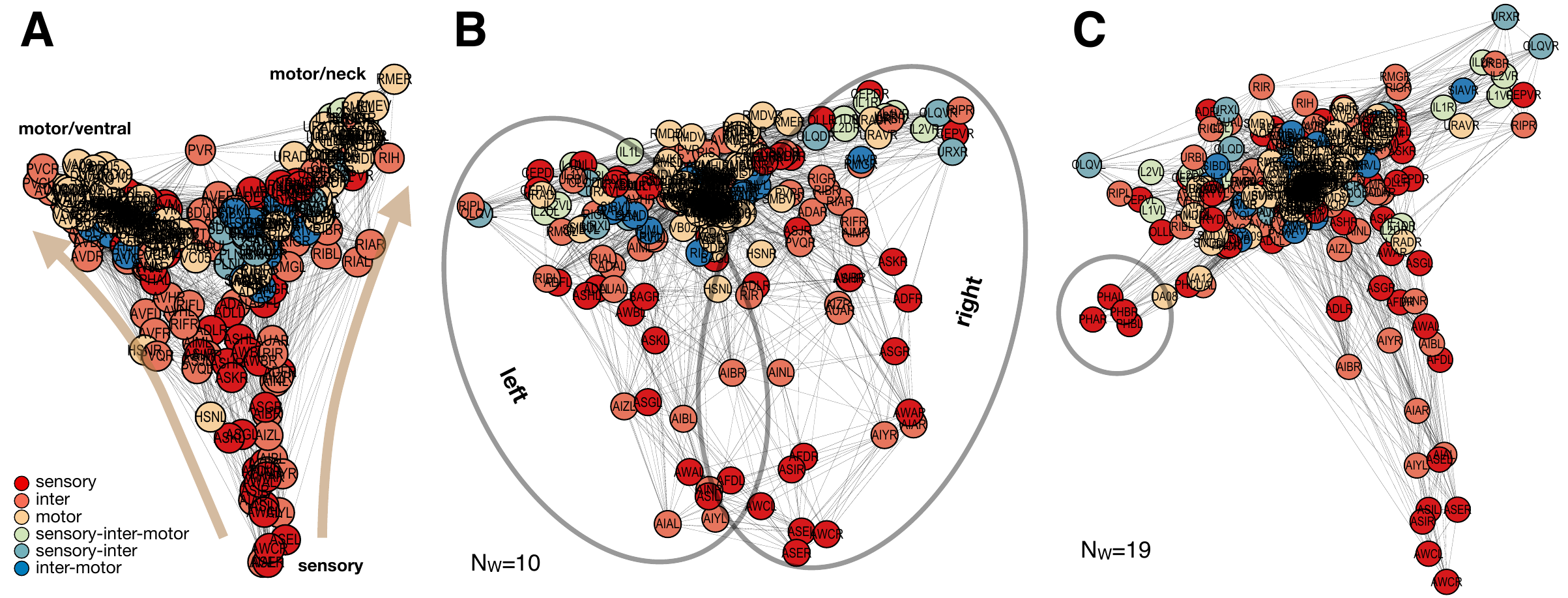}
\end{tabular}
\caption{\label{fig:celegans} (A)~Laplacian embedding on the two eigenvectors with smallest non-zero eigenvalues. (B)~Slepian embedding when selecting sensory neurons and bandwidth $N_W=10$. (C)~Slepian embedding when selecting sensory neurons and bandwidth $N_W=19$.}
\end{sidewaysfigure}

In their modeling work, Varshney et al~\cite{Varshney.2011} studied network properties of the worm connectome using different approaches, including Laplacian embedding. In particular, the topological view generated by mapping on the first two eigenvectors with smallest non-zero eigenvalues already reveals an interesting organization, see Fig.~\ref{fig:celegans}A. The horizontal dimension ($\vc u_2$) reveals two lobes of neurons mostly according to ventral and neck neurons, respectively. The vertical dimension ($\vc u_3$) show mostly a sensory-inter-motor neuron gradient (from bottom to top), which is very similar to previous results~\cite{Varshney.2011}. 

\subsection{Graph Slepian Analysis}
To provide a 2-D visualization, we want to maximize the variance captured by the Slepians in the selected subgraph and avoid the selection of the ``most important'' Slepian vectors. Therefore, we construct the restricted Gram matrix that contains the inner products of the graph Slepians computed on the subgraph $\mathcal{S}$; i.e., we have $\ma K = \ma G^T \ma S \ma G$. We then diagonalize this matrix using the SVD as $\ma K=\ma V \ma \Lambda^{(\ma K)} \ma V^T$ and project $\ma G$ onto this basis to obtain coordinates as $\ma G\ma V$, from which we keep the two dimensions with largest eigenvalues in $\ma \Lambda^{(\ma K)}$. The columns of $\ma V$ can be considered as a data-driven spectral window that allows maximizing the variance in the visualization for nodes that are part of $\mathcal{S}$. 

We specifically pick the subgraph $\mathcal{S}$ as the sensory neurons, or polymodal neurons with a sensory component. A total of 83 neurons are selected, which represents $29.75\%$ of all neurons. For low bandwidth up to $N_W=5$, the Slepian view on the network is basically identical to the Laplacian embedding. Then, for higher bandwidth such as $N_W=10$, the energy concentration in the subset of sensory neurons is able to increase and a different structure is revealed, as shown in Fig.~\ref{fig:celegans}B. First, many of the motorneurons have moved towards the center of the representation, which means that many of them are not so well connected to sensory neurons. Second, the two lobes of ventral/neck neurons are replaced by two lobes of mainly left/right neurons, including some remaining inter- and motorneurons. This structure is preserved as the bandwidth is further increased, although some additional organization within $\mathcal{S}$ is uncovered, such as the branch of pharynx neurons for $N_W=19$, see Fig.~\ref{fig:celegans}C.

For these results, we used the graph Slepian design based on the modified embedded distance, however, the graph Slepian design based on the energy concentration leads to almost identical results. The reason is that the summarizing method described before relies on the subspace spanned by the Slepians and not on their eigenvalues, and these spaces are very similar. However, for any spectral graph operations (e.g., filtering~\cite{Shuman.2013} or wavelet construction~\cite{Hammond.2011}), the Slepian design based on the modified embedded distance is advantageously as the eigenvalues can be used as frequencies; for an exemple see~\cite{vandeville1701}. 

\begin{figure}[t]
\centering
\begin{tabular}{cc}
(A) & (B) \\
\includegraphics[width=7cm]{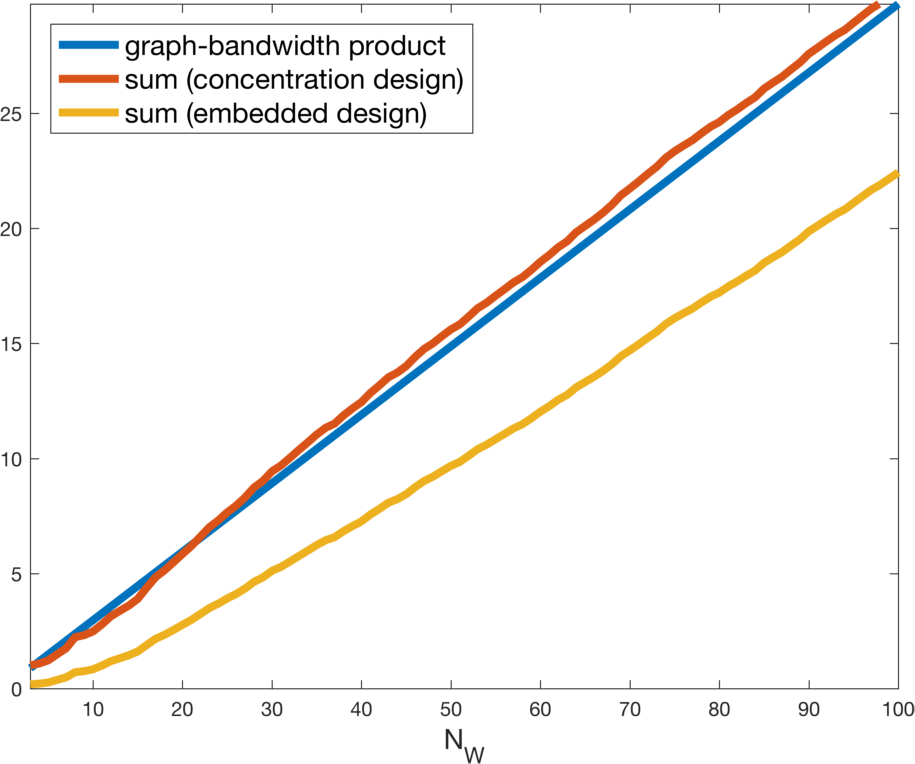} & 
\includegraphics[width=7cm]{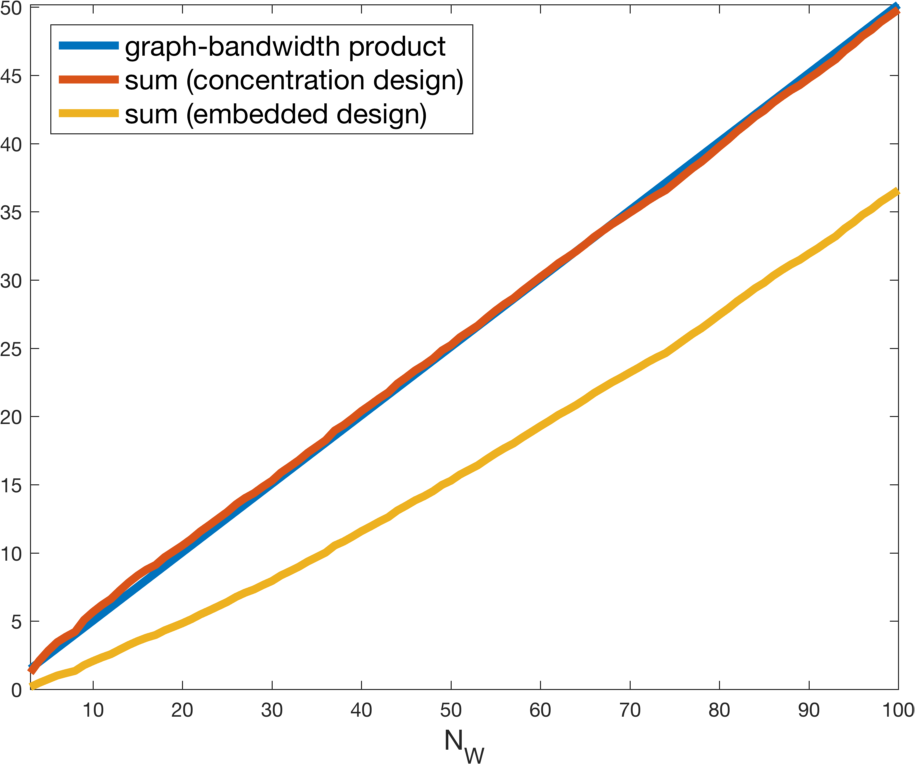} 
\end{tabular}
\caption{\label{fig:dim} Plots of the graph-bandwidth product and the sum of concentration values for the two Slepian designs and varying bandwidth $N_W$ when selecting (A)~sensory or (B)~motor neurons.}
\end{figure}

One well known phenomenon for classical Slepian functions is the phase transition between concentrated and non-concentrated functions, which is given by the Shannon number or time-bandwidth product. For the graph Slepians, the graph-bandwidth product corresponds to $K=N_W N_S / N$. However, as opposed to the conventional case, this is not necessarily equal to the sum of concentration values $\sum_{k=1}^{N_W} \mu_k$ due to the irregularity of the domain. In Fig.~\ref{fig:dim}A, we plot the graph-bandwidth product and the sum of concentration values for the two Slepian designs when selecting the sensory neurons. We notice that for the concentration design, the sum is almost identical to $K$ as the bandwidth varies. For the embedded-distance design, we notice a gap between both measures, although the sum of concentration values still increases approximately linearly. In Fig.~\ref{fig:dim}B, we show a similar behavior when selecting the motor neurons (140 in total). 

\section{Conclusion}
We have shown how graph Slepians can be designed and used to guide network analysis. The proposed method is flexible as it allows to easily recompute the Slepian vectors for different choices of the subgraph and bandwidth, once the (truncated) eigendecomposition of the graph Laplacian is available. For large(r) graphs, one can rely on efficient large-scale solvers~\cite{Lehoucq.1996} implemented in widely available software libraries such as ARPACK. Future extensions of this approach could include the relationship with graph uncertainty principles~\cite{Agaskar.2013,Tsitsvero.2016,Teke.2017}, statistical resampling for graphs~\cite{pirondini1601}, or discovery of hierarchical graph structure~\cite{Arenas.2008, Irion.2014} by gradual refinement of the subgraph. The design could also be extended for directed graphs using recent extensions of spectral decompositions for directed graphs~\cite{Sandryhaila.2013,Mhaskar.2016}.

The Matlab implementation of the graph Slepians is publicly available\footnote{https://codeocean.com/2017/05/10/graph-slepians}.


\acknowledgements
\label{sec:ack}
This work was supported in part by the Swiss National Science Foundation, and in part by the Center for Biomedical Imaging (CIBM) of the Geneva - Lausanne Universities and the EPFL, the foundations Leenaards and Louis-Jeantet.


\end{document}